# Time to Embrace Natural Language Processing (NLP)-based Digital Pathology: Benchmarking NLP- and Convolutional Neural Network-based Deep Learning Pipelines

Min Cen[1], Xingyu Li[2], Bangwei Guo[1], Jitendra Jonnagaddala[3*], Hong Zhang[2*], Xu Steven Xu[4*]

[1]School of Data Science, University of Science and Technology of China
[2]Department of Statistics and Finance, School of Management, University of Science and Technology of China
[3]School of Population Health, UNSW Sydney, Kensington, NSW, Australia
[4]Clinical Pharmacology and Quantitative Science, Genmab Inc., Princeton, New Jersey, USA
*Corresponding author. jitendra.jonnagaddala@unsw.edu.au (Jitendra Jonnagaddala), zhangh@ustc.edu.cn (Hong Zhang), sxu@genmab.com (Xu Steven Xu)

**Conflict of interest statements**：XSX is an employee of Genmab, Inc. Genmab did not provide any funding for this study.

Main text: 2988 words.

# Abstract

Natural language processing (NLP)-based computer vision models, particularly vision transformers, have been shown to outperform convolutional neural network (CNN) models in many imaging tasks. However, most digital pathology artificial-intelligence models are based on CNN architectures, probably owing to a lack of data regarding NLP models for pathology images. In this study, we developed digital pathology pipelines to benchmark the five most recently proposed NLP models (vision transformer (ViT), Swin Transformer, MobileViT, CMT, and Sequencer2D) and four popular CNN models (ResNet18, ResNet50, MobileNetV2, and EfficientNet) to predict biomarkers in colorectal cancer (microsatellite instability, CpG island methylator phenotype, and *BRAF* mutation). Hematoxylin and eosin-stained whole-slide images from Molecular and Cellular Oncology and The Cancer Genome Atlas were used as training and external validation datasets, respectively. Cross-study external validations revealed that the NLP-based models significantly outperformed the CNN-based models in biomarker prediction tasks, improving the overall prediction and precision up to approximately 10% and 26%, respectively. Notably, compared with existing models in the current literature using large training datasets, our NLP models achieved state-of-the-art predictions for all three biomarkers using a relatively small training dataset, suggesting that large training datasets are not a prerequisite for NLP models or transformers, and NLP may be more suitable for clinical studies in which small training datasets are commonly collected. The superior performance of Sequencer2D suggests that further research and innovation on both transformer and bidirectional long short-term memory architectures are warranted in the field of digital pathology. NLP models can replace classic CNN architectures and become the new workhorse backbone in the field of digital pathology.

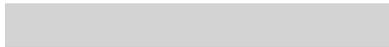

# 1. Introduction

Convolutional neural networks (CNNs), such as ResNet (He et al., 2016) and EfficientNet (Tan and Le, 2019), have served as the mainstream architecture for digital pathology, including tumor detection (Campanella et al., 2019; Pinckaers et al., 2021), subtyping (Lu et al., 2021; Wang et al., 2019; Zhu et al., 2021) and grading (Bulten et al., 2020; Shaban et al., 2020; Ström et al., 2020). CNNs have shown excellent performance in predicting molecular biomarkers using hematoxylin and eosin (H&E)-stained histopathological images. Yamashita et al. used MobileNetV2 to predict microsatellite instability (MSI) (Yamashita et al., 2021). Laleh et al.(2022) adopted EfficientNet and achieved outstanding performance in subtyping, gene mutation prediction, and biomarker prediction for four types of cancer.

Natural language processing (NLP) has been dominant in the fields of speech recognition (Dong et al., 2018), synthesis (Li et al., 2019), text to speech translation (Vila et al., 2018), and natural language generation (Topal et al., 2021). The emergence of transformers (Vaswani et al., 2017) has started a new era of NLP. Inspired by the success of transformers, Dosovitskiy et al. (2020) proposed the vision transformer (ViT) for image analysis. Subsequently, various variants of the ViT have emerged and achieved state-of-the-art results, outperforming CNNs (Liu et al., 2021; Mehta and Rastegari, 2021; Zhai et al., 2022). Recently, Tatsunami and Taki (2022) proposed a sequencer using long short-term memory (LSTM), another type of NLP, and achieved competitive results with ViT on ImageNet classification and recognition tasks.

However, NLP-based models are scarce in the field of digital pathology. Laleh et al. (2022) demonstrated that a ViT model outperforms some CNN architectures in different digital pathology tasks. Despite the perception that large training datasets are required for transformers, our recent study demonstrated that Swin-T transformers achieve superior predictive performance in predicting microsatellite instability (MSI) and other biomarkers in colorectal cancer and are more robust than CNN models for small training sets (Guo et al., 2022a). However, Deininger et al. (2022) indicated that ViT models exhibit similar performances to ResNet18 but require more effort in training. Given the potential of NLP-based methods, although limited in terms of application in the current literature, the most recently developed NLP-based models must be fully and urgently investigated and compared in a clinical study setting (i.e., with only relatively small datasets available) to provide proper recommendations and guidance for digital pathology.

In this study, we aimed to develop digital pathology pipelines to benchmark the five most recently proposed NLP models (ViT (Dosovitskiy et al., 2020), Swin Transformer (Swin-T) (Liu et al., 2021), MobileViT (Mehta and Rastegari, 2021), CMT (Guo et al., 2022b), and Sequencer2D (Tatsunami and Taki, 2022)) and four popular CNN models (ResNet18(He et al., 2016), ResNet50 (He et al., 2016), MobileNetV2 (Sandler et al., 2018), and EfficientNet (Tan and Le, 2019)) to predict key molecular pathways and gene mutations (i.e., MSI, *BRAF* mutation, and CpG island methylator phenotype [CIMP]) in colorectal cancer (CRC) using H&E-stained whole-slide images (WSIs). This study provides data to better understand the performance of advanced NLP models compared with that of popular CNN models in pathology prediction tasks and recommendations regarding preferred models in digital pathology. To the best of our knowledge, this is the first study to apply and evaluate advanced NLP models such as CMT, MobileViT, and Sequencer2D in digital pathology prediction tasks.

## 2. Methods

We developed five NLP-based and four CNN-based pipelines to predict molecular-level information, including the MSI status, *BRAF* mutation, and CIMP status. The prediction pipeline (Figure 1) comprises four modules: (a) image preprocessing, (b) tumor tissue selection, (c) tile-level score prediction using NLP- or CNN-based architectures, and (d) tile-level score aggregation.

We compared the prediction performance model sizes and run times for all models.

## 2.1. Datasets

The H&E-stained WSIs of patients with CRC from two study cohorts were used: Molecular and Cellular Oncology (MCO) and The Cancer Genome Atlas (TCGA). The MCO-CRC (Ward, 2015 #83) (Jonnagaddala et al., 2016) dataset consisted of patients who underwent curative resection for CRC between 1994 and 2010 in New South Wales, Australia, and is available through the SREDH Consortium (www.sredhconsortium.org, accessed on February 15, 2023. The ground-truth labels for the MSI status of the MCO dataset were determined using mismatch repair immunohistochemistry (IHC). *BRAF* mutations (V600E) were identified via IHC, PCR, or Sanger sequencing. The TCGA dataset (publicly available at https://portal.gdc.cancer.gov) is a multicenter study of patients with stage I–IV disease, primarily from the United States. Data from TCGA-COAD and TCGA-READ cohorts were used (hereafter referred to as TCGA-CRC-DX). The ground-truth labels of the MSI status, *BRAF* mutation, and CIMP status for TCGA-CRC-DX were obtained from Liu et al. (2018) Furthermore, for a better comparison with the existing literature, we used the same TCGA-CRC-DX sub-datasets for each biomarker as those used and published by Kather et al. (2020) and Bilal et al. (2021) (referred to hereafter as TCGA-CRC-DX). For patients with more than one available WSI, one slide was randomly selected for each patient. Supplementary Table 1 lists the number of the patients' WSIs for each label that we used for training and testing.

We selected the tiles of the tumor region using a nine-class tissue classifier based on the publicly available NCT-CRC-HE-100K and CRC-VAL-HE-7K (Guo et al., 2022a) datasets. Up to 500 tumor tiles were randomly selected per patient as tiles of interest for the downstream deep learning.

### 2.2. Image Preprocessing

Following a procedure similar to that of Laleh et al. (2022), all WSIs were tessellated into small image tiles of $512 \times 512$ pixels at a resolution of 0.5 µm. The tiles were color normalized using the Macenko method (Macenko et al., 2009). Image tiles containing background or blurry images were automatically removed from the dataset during this process by the detected edge quantity (canny edge detection in Python's OpenCV package) (https://github.com/KatherLab/preProcessing). The image tiles were then resized to $224 \times 224$ pixels to fit the input of the networks.

### 2.3. NLP-Based Deep Learning Framework

- **Vision Transformer.** ViT (Dosovitskiy et al., 2020) was proposed in 2020. It can account for long-range dependencies in images and incorporate more global information than CNNs. ViT divides an input image into multiple patches ($16 \times 16$) and projects each patch into a fixed-length vector. ViT has achieved state-of-the-art results for a variety of computer vision tasks(Dosovitskiy et al., 2020). ViT was first adopted by Laleh et al. (2022) in digital pathology. The limitations of ViT include its large model size and high computing power requirements (Dosovitskiy et al., 2020).

- **Swin Transformer.** Unlike word tokens, which are the basic elements of processing in language transformers, visual elements can vary significantly in scale. Additionally, pixel resolutions in an image are much higher than those in text paragraphs. Therefore, Liu et al. (2021) proposed the Swin-T for constructing a hierarchical representation by gradually merging adjacent patches in deeper transformer layers. This allows Swin-T to easily utilize dense prediction and reduce the computational complexity from quadratic (ViT) to linear. We have shown that Swin-T can achieve excellent biomarker prediction using relatively small datasets (Guo et al., 2022a).

- **MobileViT.** Owing to the demand for lightweight networks in mobile vision tasks, the transformer-based model MobileViT (Mehta and Rastegari, 2021) was proposed by Apple Inc. in 2021 to leverage the light weight of CNNs and global representation of ViT. MobileViT primarily comprises ordinary convolution, inverted residual blocks in MobileNetV2, MobileViT blocks, global pooling, and fully connected layers. A MobileViT block preserves the order of the patches and of each pixel in each patch. Moreover, MobileViT replaces matrix multiplication with a transformer to obtain global representations and has convolutional features.

- **CMT.** CMT (Guo et al., 2022b) is a hybrid network leveraging the advantages of both CNNs and transformers. It introduces a convolution operation for fine-grained feature extraction and employs a unique module hierarchical extraction of local and global features. Most transformer-based models use a large convolution (such as the $16 \times 16$ convolution kernel in ViT) to cut the input image directly into non-overlapping tiles. However, this approach loses the 2D spatial features, edge information, and many details in the patch.

Therefore, CMT uses a structure formed by stacking multiple 3 × 3 convolutions to down sample and extract detailed features. CMT has achieved state-of-the-art accuracy in image classification (Guo et al., 2022b).

- **Sequencer2D.** Sequencer2D (Tatsunami and Taki, 2022) uses LSTMs rather than self-attention layers, such as transformers, because an LSTM is more memory-economical and parameter-saving than a transformer. Sequencer2D takes non-overlapping tiles as input and processes them to a feature map; this feature map is then fed to bidirectional long short-term memory (BiLSTM) layers and a multi-layer perceptron with a linear classifier by a global average pooling layer at the top of the model. The BiLSTM architecture allows the vertical and horizontal axes to be parallel, which improves the accuracy and efficiency of the sequencer owing to the reduced sequence length, and yields a spatially meaningful receptive field. Sequencer2D has achieved competitive performance in image classification (Tatsunami and Taki, 2022).

The Supplementary Materials and Methods section describes the CNN models.

## 2.4. Training, Testing and Hyperparameter Tuning

All NLP- and CNN-based models were trained on the MCO-CRC dataset. Five-fold cross-validations were performed to select the optimal model. The models selected were then externally tested on the TCGA-CRC-DX dataset. Weighted cross-entropy loss and Adam optimizer were used to train the models. The initial learning rate of the models was set to 0.0001. Early stopping was used to avoid overfitting during training.

**2.5. Statistical Analyses**

The predictive performance of the deep learning models was evaluated using the values of the area under the receiver operating curve (AUROC) to gauge the overall predictive performance and that of the area under the precision-recall curve (AUPRC) to gauge the precision. Bootstrapping (1,000×) was used to calculate the 95% confidence intervals (Cis) of the AUROC and AUPRC values. To compare the model run times, the training time for an epoch and prediction time (all patients in the TCGA-CRC-DX) for the MSI status were recorded.

## 3. Results

**3.1. Predictive Performance**

*MSI*

All NLP-based methods outperformed the CNN-based algorithms in predicting the MSI status regarding both the AUROC and AUPRC (Tables 1/2 and Figure 2). Among the NLP-based models, the MobileViT model achieved the lowest AUROC and AUPRC values (82.9% and 55.3%, respectively); however, these were higher than the highest AUROC (80.9%, ResNet50) and AUPRC (43.9%, MobilNetV2) values obtained by the CNN models tested. The CMT achieved a state-of-the-art performance with an AUROC of 90.6% (with a 95% CI of 86.2–94.4%), which is almost 10% higher than that of the best-performing CNN-based model, that is, ResNet50 with an AUROC value of 80.9% (95% CI: 75.6–85.9%). In addition, the Swin-T, ViT, Sequencer2D and MobileViT models demonstrated competitive AUROC values in the MSI prediction, i.e., 86.8% (95% CI:80.8–92.1%), 86.3% (95% CI: 80.1–91.7%), 84.3% (95%

CI: 77.9–90.1%), and 82.9% (95% CI: 76.5–88.8%), respectively. The Swin-T model for the MSI produced the highest AUPRC value (69.7%), almost 26% higher than that of the best-performing CNN model (MobilNetV2; AUPRC = 43.9%). The CMT, ViT, and Sequencer2D models achieved similarly high AUPRC values, ranging from 65.6% to 68.4%.

*BRAF mutation*

Overall, the NLP-based models predicted *BRAF* mutations better than the CNN-based models. The Sequencer2D model achieved the highest AUROC (80.2%; 95% CI: 73.4–86.0%), followed by ViT (78.6%) and Swin-T (78.4%). The CMT model achieved the lowest AUROC (76.2%) among the NLP-based models. For the CNN-based methods, ResNet50 produced an AUROC (77.3%), which is on par with that of MobileViT.

In terms of the AUPRC, Swin-T achieved the best precision (AUPRC = 43.2%; 95% CI: 30.8–55.8%). Sequencer2D also achieved an outstanding AUPRC of 40.2% (95% CI: 29.3–55.3%), which was higher than that achieved by any of the CNN models tested. However, it should be noted that, compared with Swin-T and Sequencer2D, the MobileViT, CMT, and ViT models produced suboptimal precisions, with an AUPRC of approximately 35%.

*CIMP status*

The NLP models predicted the CIMP status better than the CNN models. Sequencer2D and Swin-T were the two best models for predicting CIMP, achieving AUROC values of 79.5% (95% CI: 71.9–86.0%) and 78.2% (95% CI: 70.3–85.0%), respectively. In contrast, the best

AUROC value yielded by a CNN-based model was 72.5% (95% CI: 64.4–80.4%) by EfficientNet.

In addition, Sequencer2D achieved the highest precision in predicting the CIMP status, with an AUPRC of 60.3% (95% CI: 48.2%–72.7%). The Swin-T and ViT models also achieved excellent precision results (AUPRC = 53.7% and 53.9%, respectively). In contrast, the highest AUPRC achieved by a CNN-based model (45.6%; 95% CI: 32.6–59.3%; MobileNetV2) was 14.7% lower than that of Sequencer2D.

## 3.2. Relationships Between the Model Complexity, Predictive Performance, and Run Time

The sizes of the CNN-based models were generally smaller than those of the NLP-based models (Figure 3). MobileNetV2, EfficientNet, and MobileViT are the most lightweight models (~4M parameters), whereas the ViT model has the most model parameters (85.8M). The model sizes of CMT and Sequencer2D are comparable to that of ResNet50 (approximately 25M). Swin-T has approximately 48.84M parameters, which is substantially lower than that of ViT.

As expected, the model predictive performance (AUROC and AUPRC) generally correlated with the model complexity and size (Figure 3). The CMT and Sequencer2D models exhibited outstanding predictive performance, despite their relatively small sizes. Similarly, Swin-T consistently outperformed the largest and most complex tested model ViT and tended to provide an excellent precision.

A positive relationship was observed between the model size and training/prediction times (Figure 4). As a tradeoff between their better predictive performances, the NLP models generally require longer training and prediction times compared with those of the CNN models. Interestingly, the CMT model required a much longer training time than the larger ViT model, suggesting a higher level of model complexity compared to the other models. Similar training and prediction times were required for the Sequencer2D, Swin-T, and ViT models, despite the differences in their model sizes. Overall, Sequencer2D and Swin-T exhibited the highest efficiencies (i.e., the highest performance/complexity ratio) among all CNN- and NLP-based models tested.

4. **Discussion**

H&E-stained WSIs are ubiquitously available for almost all cancer patients. Advances in artificial intelligence have led to significant progress in digital pathology analysis using H&E-stained WSIs, including the prediction of molecular characteristics, pathways, and clinical outcomes in cancer patients (Bilal et al., 2021; Echle et al., 2022; Ilse et al., 2018; Kather et al., 2019; Li et al., 2022a; Li et al., 2022b; Lu et al., 2021; Schmauch et al., 2020). However, to date, most digital pathology pipelines have been based on CNN architectures (Echle et al., 2020; Echle et al., 2022) despite the recent emergence of NLP-based models. NLP architectures are more flexible for learning long-range interactions in images, whereas CNN networks tend to be more rigid and focused on local features from nearby pixels (Dosovitskiy et al., 2020). NLP-based models, such as transformers, have been demonstrated to outperform CNNs in many computer vision tasks (Dosovitskiy et al., 2020; Liu et al., 2021; Tatsunami and Taki, 2022).

In this study, we benchmarked five advanced NLP models and four popular CNN models in terms of predicting the MSI status, *BRAF* mutation, and CIMP status in CRC patients. The external validation on the TCGA-CRC-DX dataset demonstrated that the NLP-based models significantly outperformed the CNN-based models. In particular, compared with the best-performing CNN-based model, the NLP-based models improved the AUROC in predicting the MSI status, *BRAF* mutation, and CIMP status by 9.7%, 2.9%, and 7.0%, respectively, while improving the AUPRC (precision) by 25.8%, 1.2%, and 14.7%, respectively. Furthermore, the Sequencer2D and Swin-T models exhibited a remarkable robustness for precision, providing the highest AUPRC values for all three biomarkers.

The MSI status is critical in treating CRC patients, and immunotherapies such as pembrolizumab and nivolumab have been approved by health authorities to treat CRC patients with MSI-High (André et al., 2020; Marcus et al., 2019). In predicting the MSI status, the performance achieved by CMT is the best among all artificial-intelligence models reported, including the most recent state-of-the-art model reported by Echle et al. (2022) trained on a large, multicenter dataset (N = 7917; AUROC = 91%), which is much larger than the MCO-CRC dataset (N = 1138) used for training in this study. The current clinical gold-standard test for MSI is based on IHC (Hampel et al., 2008; Stjepanovic et al., 2019). We showed that the NLP models, particularly the CMT model, can serve as a testing tool for the MSI status, reduce the turnaround time, and save on expenses. In addition, the Sequencer2D model for the *BRAF* mutation and CIMP status achieved a state-of-the-art performance compared with all existing models in the current literature. Notably, the AUROC values achieved by Sequencer2D in our external, cross-cohort validation were even higher than those obtained by Bilal et al. (2021) from an intra-cohort validation (i.e., four-fold cross-validation) using the TCGA-CRC-DX dataset (AUROC = 79%).

The NLP models have certain limitations. We showed that NLP-based models are generally larger and more complex than CNN-based models; consequently, NLP models generally require longer training and prediction times compared with CNN models. Nevertheless, the most recent Sequencer2D model successfully reduced their size to a level comparable to that of ResNet50, while consistently outperforming the larger and more complex models such as the original ViT model. Overall, Sequencer2D and Swin-T demonstrated the highest performance/complexity ratio among all CNN- and NLP-based models tested.

Finally, unlike transformers, Sequencer2D is built on a more traditional BiLSTM architecture. In our benchmarking experiments, Sequencer2D tended to exhibit similar or better prediction performance in terms of both AUROC and AUPRC compared with the transformer models, while requiring less training and prediction run times than the ViT, Swin-T, and CMT models. The success of Sequencer2D highlights the advantages of LSTM, as it may provide a novel and more feasible choice for digital pathology tasks and suggests that further research and innovation on BiLSTM architectures should be conducted in the fields of digital pathology and computer vision.

## Acknowledgements

The work of MC, XL, BG and HZ was partially supported by National Natural Science Foundation of China (No. 12171451 and No. 72091212) and Anhui Center for Applied Mathematics. JJ is funded by the Australian National Health and Medical Research Council (No. GNT1192469) who acknowledges the funding support received through the Research Technology Services at UNSW Sydney, Google Cloud Research (award# GCP19980904) and NVIDIA Academic Hardware grant programs. And the MCO_CRC dataset was provided by the SREDH Consortium's (downloaded from https//:www.sredhconsortium.org, accessed on 15 November 2022) Translational Cancer Bioinformatics working group.## Author contributions statement

XSX, MC and HZ contributed to design of the research. JJ, XSX, XL and MC contributed to data acquisition. MC, BG and XSX contributed to data analysis. XSX, MC, JJ and HZ wrote the manuscript; and all authors critically reviewed the manuscript and approved the final version.

**Table 1: Performance statistics (AUROC) for external validation for TCGA-CRC-DX dataset.** Patient-level AUROC with a 95% confidence interval obtained via bootstrapping (1,000×) calculated for nine models in predicting MSI, *BRAF* mutation, and CIMP. Green rows indicate the NLP-based models other rows indicate CNN-based models. Best results are in bold. The two rightmost columns list the model structure category and number of model parameters, respectively.

| External validation on TCGA-CRC-DX | MSI (N=425) | *BRAF* mutation (N=500) | CIMP (N=235) | Architecture | # of Params |
|---|---|---|---|---|---|
| **ResNet18** | 78.5 (72.6-84.1) | 72.7 (64.4-79.6) | 69.9 (61.3-77.5) | CNN | 11.18M |
| **ResNet50** | 80.9 (75.6-85.9) | 77.3 (69.9-83.9) | 65.0 (56.6-73.2) | CNN | 23.51M |
| **MobileNetV2** | 79.1 (72.9-85.0) | 76.6 (68.1-83.8) | 66.2 (56.8-74.7) | CNN | 2.23M |
| **EfficientNet** | 77.3 (70.7-83.8) | 74.2 (65.7-81.9) | 72.5 (64.4-80.4) | CNN | 4.01M |
| **ViT** | 86.3 (80.1-91.7) | 78.6 (71.5-84.6) | 75.2 (67.4-82.2) | Transformer | 85.8M |
| **Swin-T** | 86.8 (80.8-92.1) | 78.4 (71.4-84.7) | 78.2 (70.3-85.0) | CNN + Transformer | 48.84M |
| **MobileViT** | 82.9 (76.5-88.8) | 77.8 (71.4-83.8) | 71.8 (64.3-78.5) | CNN + Transformer | 4.94M |
| **CMT** | **90.6 (86.2-94.4)** | 76.2 (68.4-82.9) | 71.1 (63.4-77.4) | CNN + Transformer | 24.98M |
| **Sequencer2D** | 84.3 (77.9-90.1) | **80.2 (73.4-86.0)** | **79.5 (71.9-86.0)** | BiLSTM | 27.27M |

**Table 2: Performance statistics (AURPC) for external validation for TCGA-CRC-DX dataset.** Patient-level AUPRC with a 95% confidence interval obtained via bootstrapping (1,000×) calculated for nine models in predicting MSI, *BRAF* mutation, and CIMP. Green rows indicate the NLP-based models, and other rows indicate CNN-based models. Best results are in bold. The two rightmost columns list the model structure category and number of model parameters, respectively.

| External validation on TCGA-CRC-DX | MSI (N=425) | *BRAF* mutation (N=500) | CIMP (N=235) | Architecture | # of Params |
|---|---|---|---|---|---|
| **ResNet18** | 38.6 (27.8-51.5) | 30.1 (20.5-44.3) | 43.6 (32.0-58.0) | CNN | 11.18M |
| **ResNet50** | 41.8 (30.8-55.1) | 37.5 (26.8-52.0) | 37.3 (26.6-51.1) | CNN | 23.51M |
| **MobileNetV2** | 43.9 (32.0-57.1) | 39.0 (27.8-54.3) | 45.6 (32.6-59.3) | CNN | 2.23M |
| **EfficientNet** | 41.2 (30.0-55.3) | 35.8 (24.6-49.4) | 41.4 (30.7-55.8) | CNN | 4.01M |
| **ViT** | 68.4 (57.2-78.8) | 35.0 (24.9-49.7) | 53.9 (41.6-66.4) | Transformer | 85.8M |
| **Swin-T** | **69.7 (59.0-79.1)** | **43.2 (30.8-55.8)** | 53.7 (41.1-67.9) | CNN + Transformer | 48.84M |
| **MobileViT** | 55.3 (41.9-68.7) | 33.8 (24.3-47.8) | 44.7 (32.6-58.4) | CNN + Transformer | 4.94M |
| **CMT** | 66.9 (54.6-79.7) | 33.7 (23.9-46.3) | 40.0 (29.5-52.9) | CNN + Transformer | 24.98M |
| **Sequencer2D** | 65.6 (53.4-77.3) | 40.2 (29.3-55.3) | **60.3 (48.2-72.7)** | BiLSTM | 27.27M |

# Figures

**Figure 1: Pipelines for predicting** of MSI, *BRAF* mutation, and CIMP in CRC**.** MCO-CRC and TCGA-CRC-DX were used to train and test for prediction of molecular biomarkers in CRC (i.e., MSI, *BRAF* mutation, and CIMP). The whole-slide images were tessellated into non-overlapping tiles of $512 \times 512$ pixels at a resolution of 0.5 µm. The resulting tiles were then resized to $224 \times 224$ pixels and color normalized. Tumor tissues (tiles) were subsequently selected by a Swin-T-based tissue-type classifier. Up to 500 tumor tiles were randomly selected for each slide. Five NLP-based models (in orange) and four CNN-based models (in blue) were trained to predict tile-level biomarkers. Models with a red star represent models that are applied in digital pathology the first time. The predictive slide labels were obtained via tile score aggregation.

**Figure 2: Comparison of the AUROCs and AUPRCs for all models in external validation on the TCGA-CRC-DX cohort.** AUROCs and AUPRCs of the nine tested models (ResNet18, ResNet50, MobileNetV2, EfficientNet, ViT, Swin-T, MobileViT, CMT, and Sequencer2D) for prediction of MSI, *BRAF* mutation, and CIMP on the external validation dataset. Color of the circular bar represents the number of parameters.

**Figure 3: Relationships between the predictive performance and model size of all models.** AUROC and AUPRC are plotted against the number of parameters of the nine models for external validation of MSI, *BRAF* mutation, and CIMP on the TCGA-CRC-DX dataset. Color of symbols represents the model type (CNN, transformer, BiLSTM). Size of symbols represents number of parameters.

**Figure 4: Relations between training/prediction efficiency and model size of all models for MSI prediction.** Training and prediction times are plotted against number of parameters. The training time for an epoch (training set = MCO-CRC) and prediction time (for all patients in the TCGA-CRC-DX dataset) for the MSI status are recorded. Color of symbols represents the type of model (CNN, transformer, BiLSTM). Size of symbols represents number of parameters.

# Figure 1.

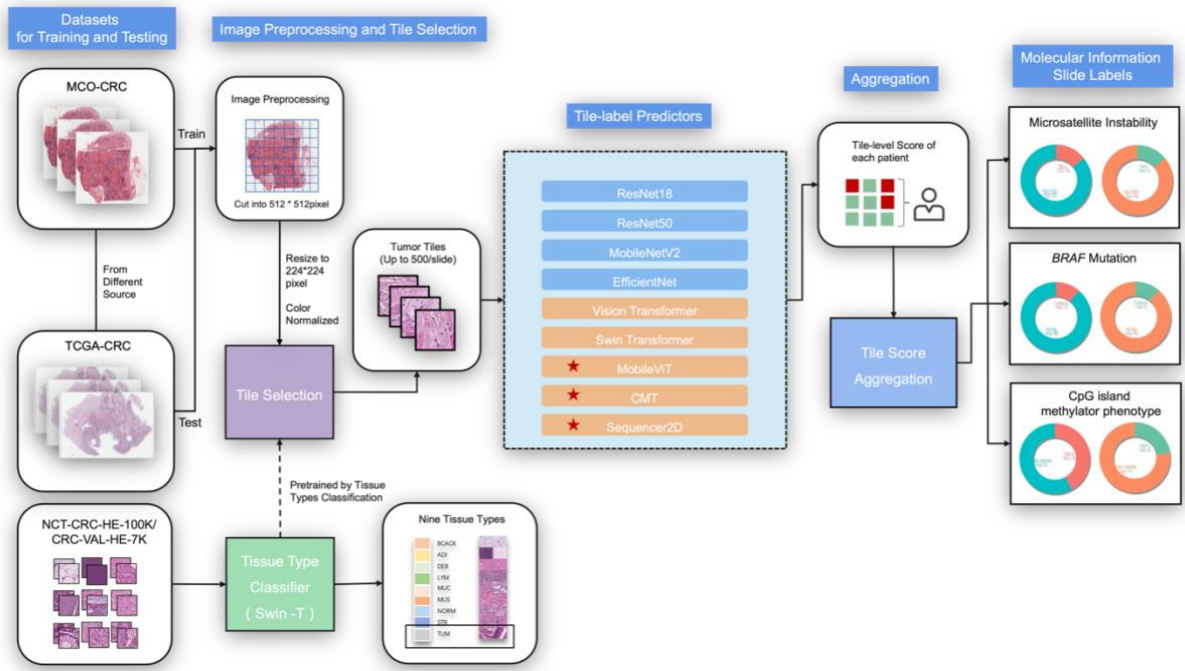

**Figure 2.**

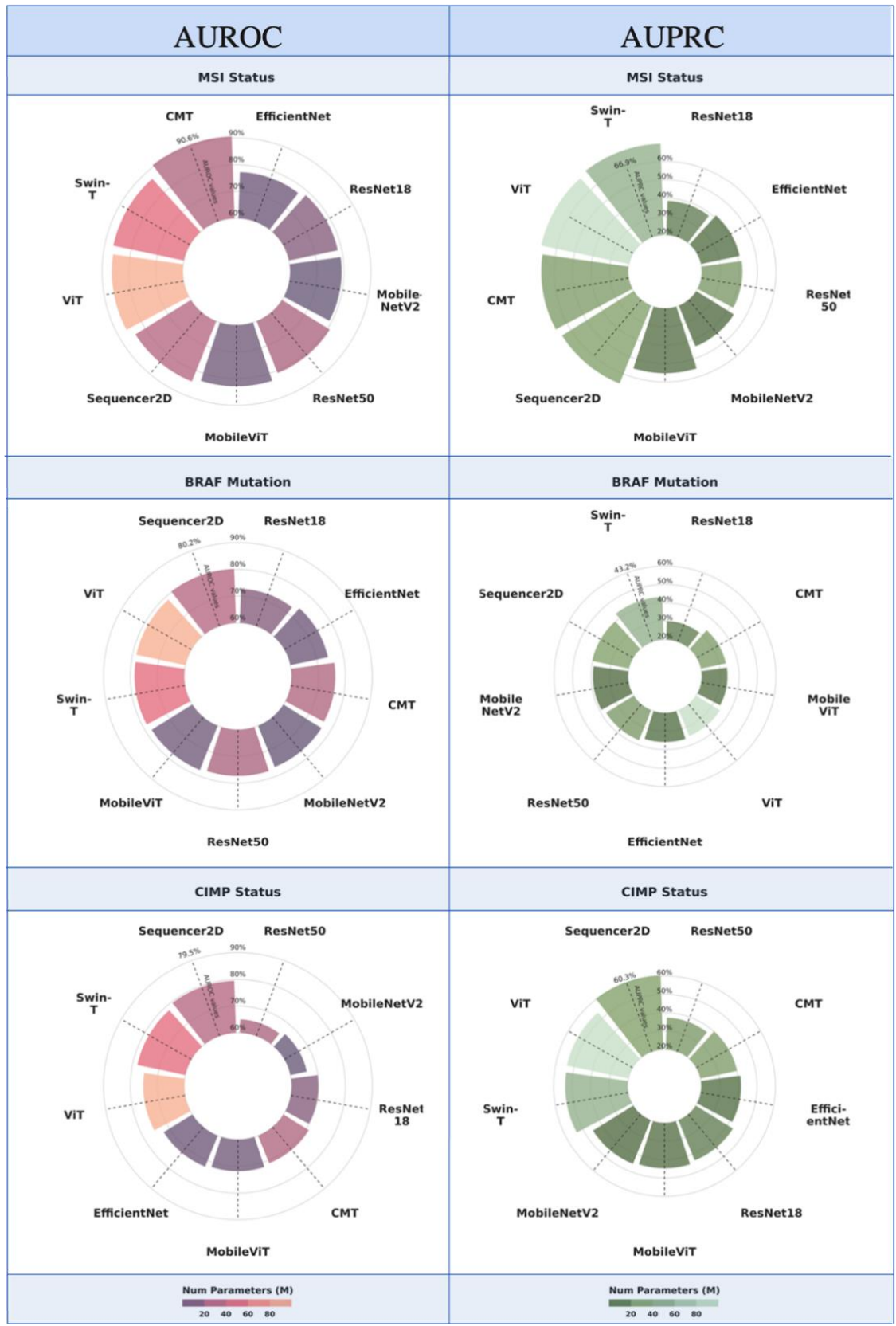

**Figure 3.**

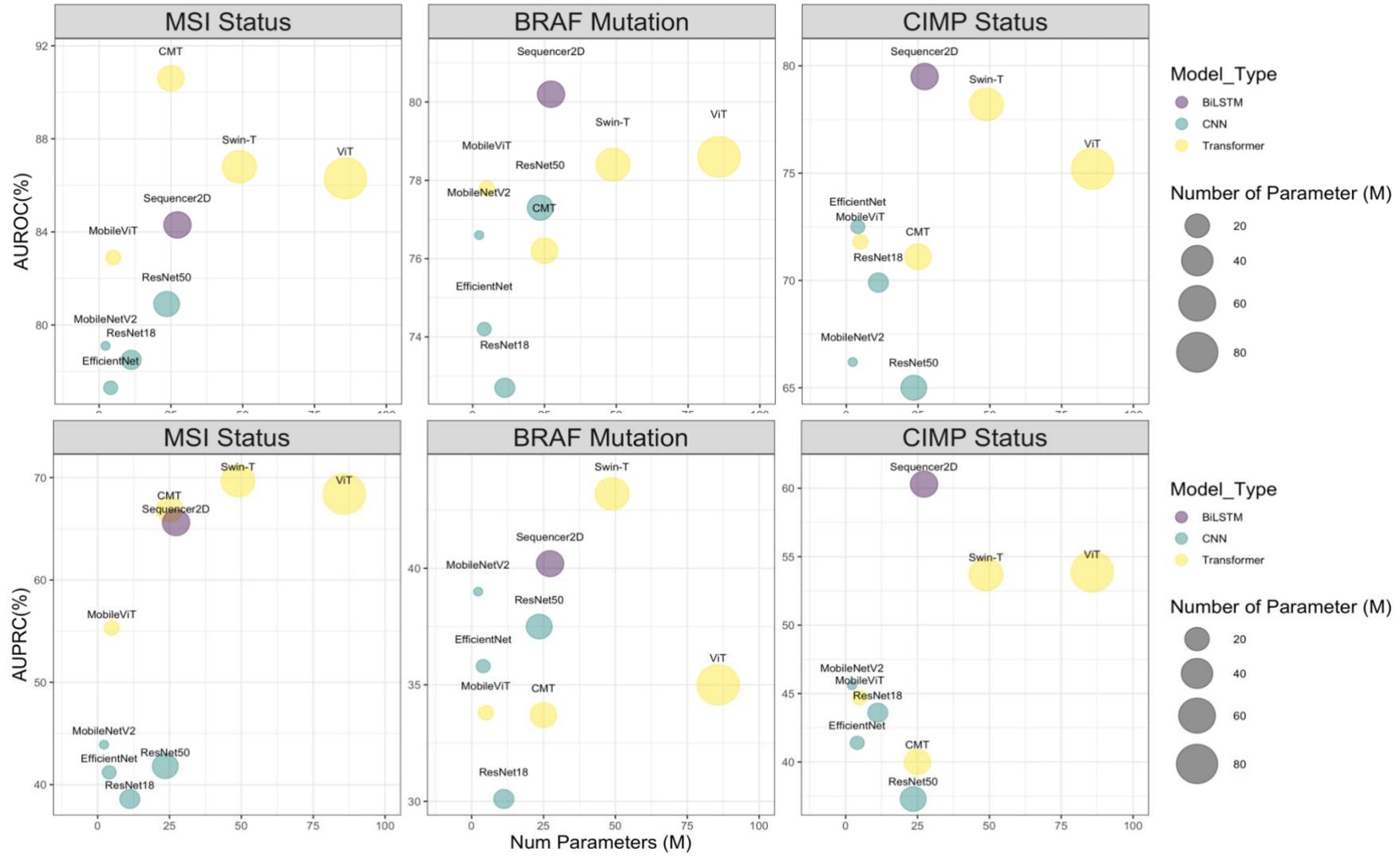

**Figure 4.**

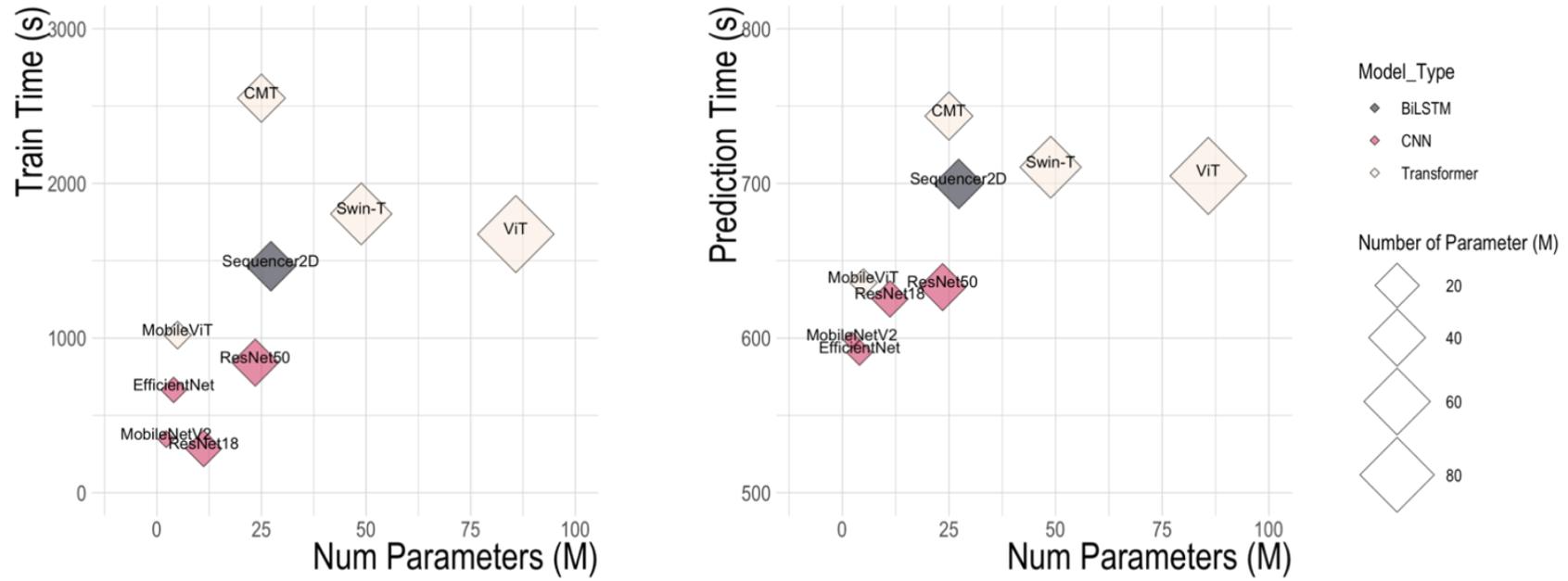

# Time to Embrace Natural Language Processing (NLP)-based Digital Pathology: Benchmarking NLP- and Convolutional Neural Network-based Deep Learning Pipelines

Supplementary Materials and Methods
Supplementary Tables
Supplementary Figures

## Supplementary Methods

**Convolutional neural network (CNN)-based deep learning models**

- **ResNet18 and ResNet50.** Residual networks (ResNet) (He et al., 2016) are among the most representative and widely used CNNs in the field of digital pathology (Echle et al., 2022; Hekler et al., 2020; Reenadevi et al., 2021). A residual block is implemented using a shortcut connection, through which the input and output of the block are added. This operation can significantly increase the training speed of the model and improve the training effect without introducing additional parameters or calculations into the network. Furthermore, the residual block can effectively solve the degradation problem in deep networks. In our study, we used ResNet18 and ResNet50 to construct our prediction pipelines.

- **MobileNetV2.** MobileNetV2 (Sandler et al., 2018) is a lightweight model based on an inverted residual structure. The main characteristic of this structure is that its input and output are thin bottleneck layers, unlike in traditional residual models (He et al., 2016). Traditional nonlinear modules are replaced with lightweight depth-wise convolutions in the intermediate expansion layer to prevent nonlinearities and build more efficient models. In 2021, Yamashita et al. (2021) adopted MobileNetV2 to predict microsatellite instability in patients with colorectal cancer (CRC) in digital pathology; it demonstrated excellent performance in a cross-cohort validation experiment.

- **EfficientNet.** EfficientNet (Tan and Le, 2019) was proposed in 2019 to improve the balance between the network depth, width, and resolution. EfficientNet is based on mobile applications and is similar to MnasNet, a multi-objective neural network architecture that searches for both accuracy and FLOPS. The main building block of EfficientNet is the mobile inversion bottleneck (MBConv). Laleh et al. (2022) used the EfficientNet series.

## Supplementary Tables

**Table S1: Number of entire side images of CRC patients regarding molecular-level information prediction.** For all biomarkers (microsatellite instability [MSI], *BRAF* mutation, and CpG island methylator phenotype [CIMP]), the number of entire-slide images in Molecular and Cellular Oncology (MCO)-CRC (for training) and The Cancer Genome Atlas (TCGA)-CRC (for testing) are listed.

| Slide Label | MCO-CRC (training) | TCGA-CRC (testing) |
|---|---|---|
| Microsatellite instability (MSI-H vs MSI-L/MSS) | 1138 (166:972) | 425 (61:364) |
| BRAF mutation (Mutational vs Wild type) | 1026 (117:909) | 500 (57:443) |
| CpG island methylator phenotype (CIMP-H vs CIMP-L/None CRC CIMP) | 364 (153:211) | 235 (54:181) |

# Supplementary Figures

**Figure S1: MobileViT, CMT, and Sequencer2D structures for predicting tile-level biomarkers.** Concrete structures of MobileViT, CMT, and Sequencer2D are used to predict the tile-level scores. For MobileViT, tiles should be first resized to 256 × 256 pixels. The gray boxes indicate the main blocks of the three models. Blocks with a red star indicate major parts based on the NLP module.

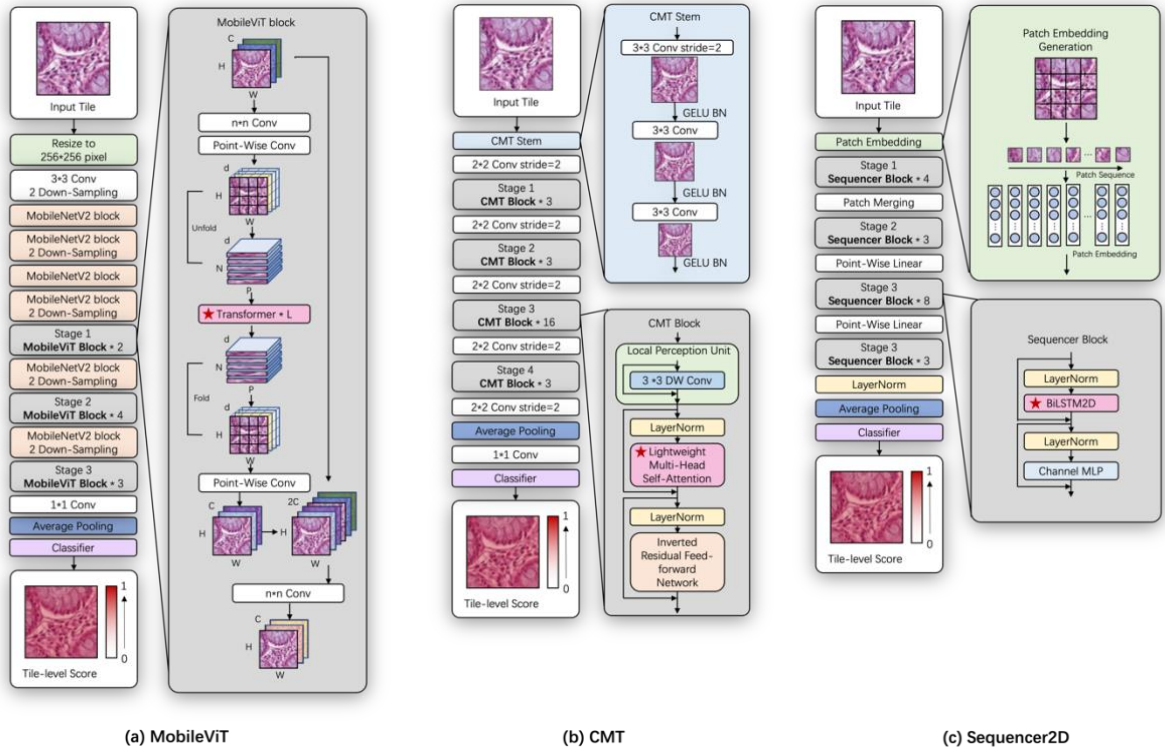

(a) MobileViT  (b) CMT  (c) Sequencer2D

**Figure S2: Predictive performances of external validation of microsatellite instability (MSI) prediction in the TCGA-CRC cohort.** Receiver operating characteristic curves (ROCs) are computed for prediction of MSI. Red-shaded areas represent the 95% confidence interval (CI) calculated via bootstrapping (1,000×). Values in the lower right of each plot indicate mean area under the receiver operating characteristic curve (AUROC; 95% CI).

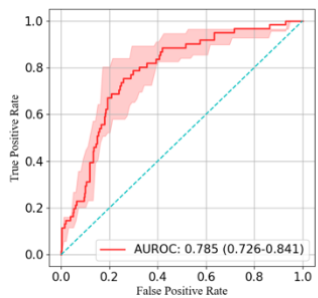
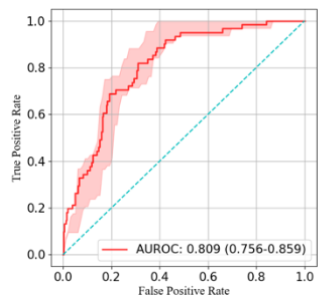
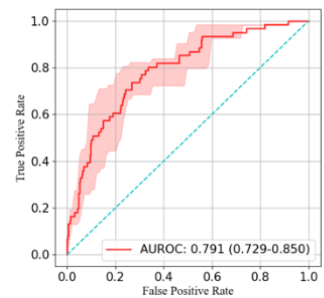

(a) ResNet18　　　　(b) ResNet50　　　　(c) MobileNetV2

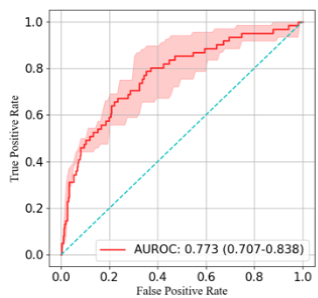
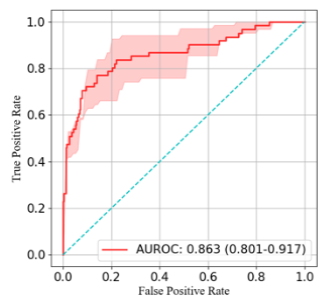
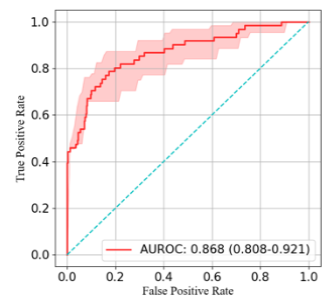

(d) EfficientNet　　　(e) ViT　　　　　　(f) Swin-T

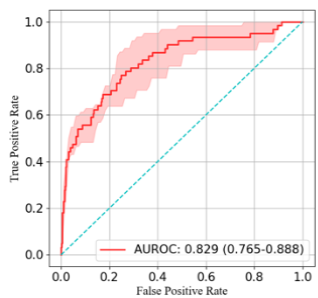
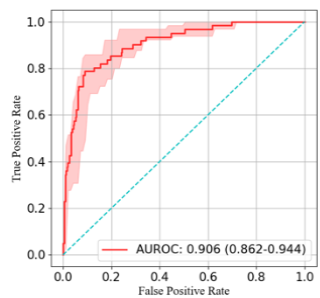
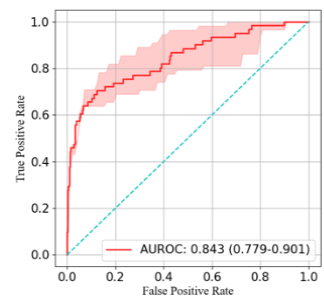

(g) MobileViT　　　　(h) CMT　　　　　(i) Sequencer2D

**Figure S3: Predictive performance of external validation of *BRAF* mutation prediction in the TCGA-CRC cohort.** ROCs are computed for prediction of *BRAF* mutation. Red-shaded areas represent the 95% CI calculated via bootstrapping (1,000×). Values in the lower right of each plot indicate mean area under the receiver operating characteristic curve (AUROC; 95% CI).

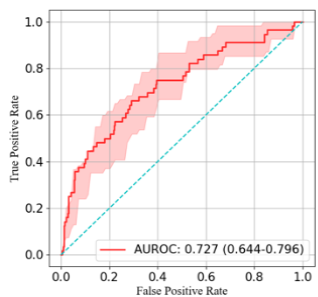 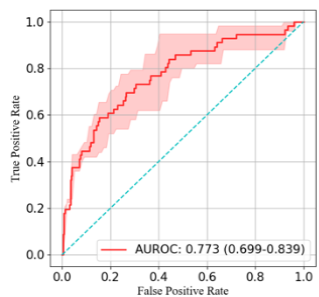 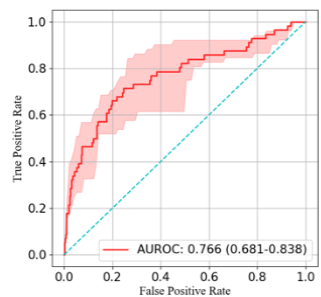

(a) ResNet18  (b) ResNet50  (c) MobileNetV2

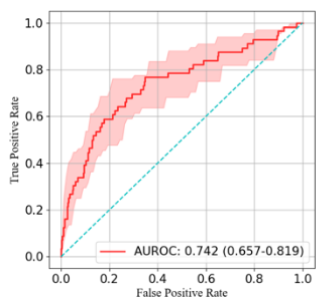 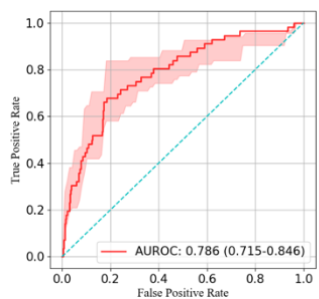 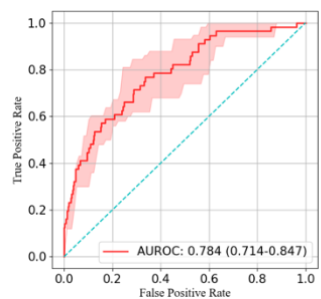

(d) EfficientNet  (e) ViT  (f) Swin-T

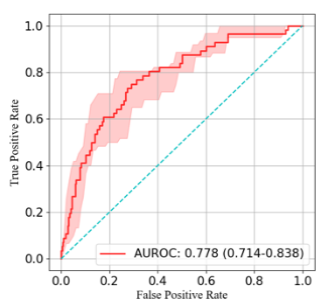 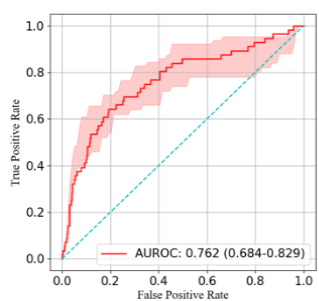 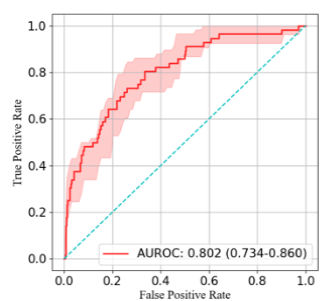

(g) MobileViT  (h) CMT  (i) Sequencer2D

**Figure S4: Predictive performance of external validation of CIMP prediction in the TCGA-CRC cohort.** ROCs are computed for prediction of CIMP. Red-shaded areas represent the 95% CI calculated via bootstrapping (1,000×). Values in the lower right of each plot indicate mean area under the receiver operating characteristic curve (AUROC; 95% CI).

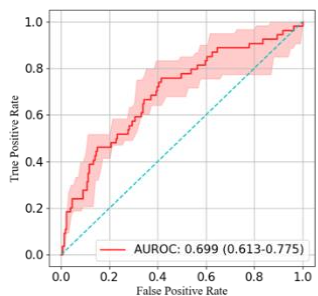
(a) ResNet18

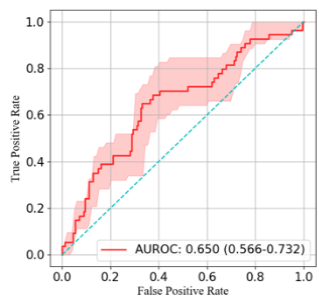
(b) ResNet50

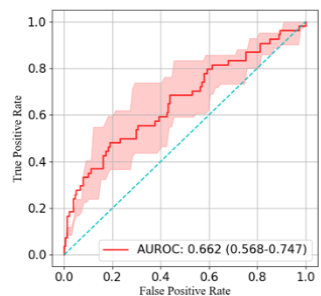
(c) MobileNetV2

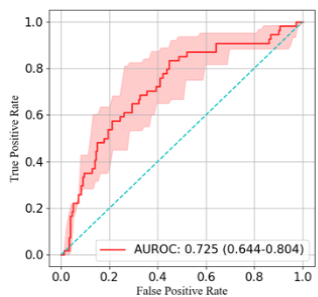
(d) EfficientNet

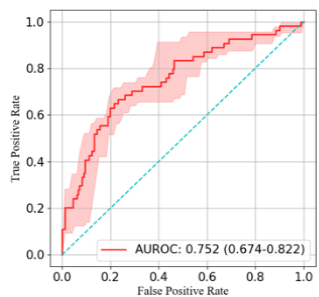
(e) ViT

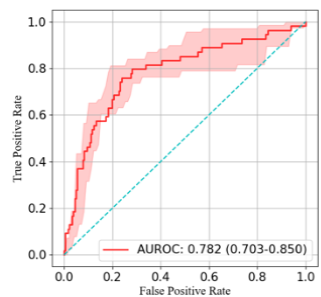
(f) Swin-T

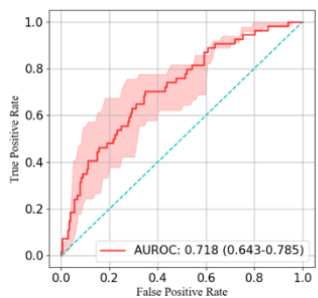
(g) MobileViT

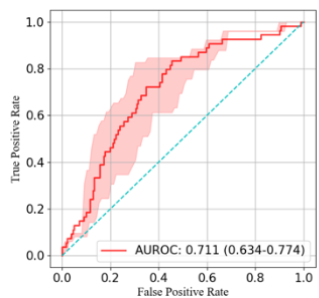
(h) CMT

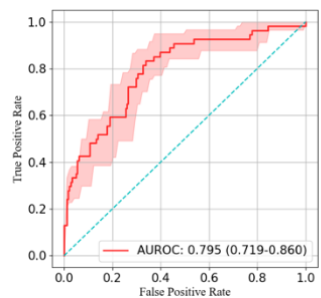
(i) Sequencer2D